\documentclass{article}

    \PassOptionsToPackage{numbers, compress}{natbib}

\usepackage[preprint]{airov24}

\usepackage{graphicx}

\usepackage[utf8]{inputenc} 
\usepackage[T1]{fontenc}    
\usepackage{hyperref}       
\usepackage{url}            
\usepackage{booktabs}       
\usepackage{amsfonts}       
\usepackage{nicefrac}       
\usepackage{microtype}      
\usepackage{xcolor}         
\usepackage{multirow}
\usepackage{amsmath}
\usepackage{tablefootnote}
\title{Improving 2D-3D Dense Correspondences with Diffusion Models for 6D Object Pose Estimation}

\author{%
  Peter Hönig \\
  Automation and Control Institute, TU Wien, Austria \\
  \texttt{hoenig@acin.tuwien.ac.at} \\
  \AND
  Stefan Thalhammer \\
  Department of Industrial Engineering, UAS Technikum Vienna, Austria \\
  \texttt{stefan.thalhammer@technikum-wien.at} \\
  \AND
  Markus Vincze \\
  Automation and Control Institute, TU Wien, Austria \\
  \texttt{vincze@acin.tuwien.ac.at} \\
}

\begin{document}

\maketitle

\begin{abstract}
    Estimating 2D-3D correspondences between RGB images and 3D space is a fundamental problem in 6D object pose estimation.
    Recent pose estimators use dense correspondence maps and Point-to-Point algorithms to estimate object poses.
    The accuracy of pose estimation depends heavily on the quality of the dense correspondence maps and their ability to withstand occlusion, clutter, and challenging material properties.
    Currently, dense correspondence maps are estimated using image-to-image translation models based on GANs, Autoencoders, or direct regression models.
    However, recent advancements in image-to-image translation have led to diffusion models being the superior choice when evaluated on benchmarking datasets.
    In this study, we compare image-to-image translation networks based on GANs and diffusion models for the downstream task of 6D object pose estimation.
    Our results demonstrate that the diffusion-based image-to-image translation model outperforms the GAN, revealing potential for further improvements in 6D object pose estimation models.

\end{abstract}

\section{Introduction}
Accurate 6D object pose estimation is essential for a range of perception tasks, such as autonomous driving, augmented reality, creating digital twins, or robotic grasping.
Numerous pose estimation methods rely on a combination of RGB and depth \cite{wang2019densefusion, bauer2020verefine}.
RGB-D sensors, which provide both color and depth data, are not always available.
Depth data is also prone to noise and other distortions, often caused by reflections of shiny, metallic, and transparent objects in the scene.
To address this issue, pose estimation from RGB images alone is considered.
State-of-the-art methods rely on estimating 2D-3D dense correspondences \cite{li2019cdpn, wang2021gdrnet, wang2019normalized, park2019pix2pose, zakharov2019dpod} between the RGB image and the 3D object model.
Altough these methods excel at inferring object poses with high visibility, they still face significant challenges posed by clutter, occlusion, image distortions, and shiny object surfaces \cite{thalhammer2023challenges}.

So far, the estimation of 2D-3D correspondences for pose estimation has been solved by image-to-image translation using direct regression \cite{li2019cdpn, wang2021gdrnet, wang2019normalized}, combinations of Generative Adversarial Networks (GANs) and U-Net architectures \cite{park2019pix2pose}, or encoder-decoder  Convolutional Neural Networks (CNNs) \cite{zakharov2019dpod}.
The abovementioned methods estimate dense correspondence maps that contain normalized XYZ object coordinates \cite{wang2019normalized}.
The 6D pose is then estimated by solving the Point-n-Point (PnP) problem via RANSAC \cite{lepetit2009epnp} or direct regression \cite{chen2022epro, chen2020end} and backpropagation.
To estimate accurate poses via PnP, the quality of the dense correspondence maps, which are estimated by the image-to-image model, is critical.

Transformer models \cite{esser2021vqgan, kim2022instaformer} are superior to traditional CNNs for image-to-image translation, when evaluated on benchmark tasks.
More recently, diffusion models \cite{rombach2022ldm} have been shown to outperform both transformer models and GANs.
However, the benchmarking tasks used to evaluate the performance of image-to-image translation do not include any tasks related to geometric feature extraction.
Therefore the question arises as to whether the performance increase of diffusion-based image-to-image translation is directly transferable to geometric feature extraction tasks.

In this paper, we conduct a comparative study to investigate the potential of diffusion models for image-to-image translation for estimating 2D-3D dense correspondences for the downstream task of 6D object pose estimation.

We summarize our contributions as follows:
\begin{itemize}
    \item We show that diffusion models are superior to GANs for estimating 2D-3D dense correspondences which leads to a higher accuracy of the subsequent 6D object pose estimation downstream task.
    \item Furthermore, our experimental findings demonstrate that diffusion models outperform GANs in the context of object segmentation. In addition, we show that diffusion models benefit more significantly from the application of online data augmentations compared to GANs.
\end{itemize}

We compare two image-to-image translation networks, namely the GAN-based Pix2Pix \cite{isola2017pix2pix} and the Brownian-Bridge Diffusion Model (BBDM) \cite{li2023bbdm} to estimate dense correspondence maps from RGB images.
To solve the subsequent PnP problem, we use RANSAC E-PnP on the dense correspondences maps.
We evaluate on the challenging Linemod-Occluded (LMO) dataset \cite{brachmann2014linemodoccluded}, which contains clutter and occlusion.

This paper is structured as follows: In Section \ref{Related Work} we give an overview of current methods for 2D-3D dense correspondence estimation.
Our method is described in Section \ref{Method} and the experimental setup is explained in Section \ref{Experimental Setup}. 
We present our results in Section \ref{Results} and conclude in Section \ref{Conclusion}.

\section{Related Work}
\label{Related Work}
In this section, we summarize the current state-of-the-art regarding types of 2D-3D correspondences and machine learning architectures for 2D-3D dense correspondences estimation for the downstream task of pose estimation.

In applications without available depth information, 6D object pose estimation has to be solved from RGB images alone.
To solve pose estimation solely from RGB images, 2D-3D correspondences have to be established.
Earlier works create these 2D-3D correspondences by utilizing (sparse) keypoints \cite{rad2017bb8, peng2019pvnet, hu2019segmentation}.
However, more recent works use dense correspondences \cite{zakharov2019dpod, park2019pix2pose, li2019cdpn, wang2021gdrnet} or surface embeddings \cite{su2022zebrapose}.
Dense correspondences are pixel-wise, as opposed to keypoints, allowing for increased resilience against occlusion or reflections \cite{thalhammer2023challenges}.
uv-coordinates were shown to be effective in \cite{li2019cdpn, zakharov2019dpod}, but were replaced with normalized object coordinate spaces (NOCS) \cite{wang2019normalized} recently \cite{wang2021gdrnet, park2019pix2pose} for further accuracy improvement. 

The core principle of 2D-3D dense correspondences estimation can be defined as an image-to-image translation task.
Given an RGB image $I_{RGB}$, the objective of the image-to-image translation function $\mathcal{F}$ is to estimate the 2D-3D dense correspondences image $I_{DC}$, as defined by the function $I_{DC} = \mathcal{F}(I_{RGB})$.
The function $\mathcal{F}$ is approximated, using a multitude of training pairs.
In order to approximate $\mathcal{F}$, various image-to-image model architectures are used.
\cite{zakharov2019dpod, li2019cdpn, wang2021gdrnet} use direct regression models with encoder-decoder structures.
\cite{park2019pix2pose} uses a U-Net architecture with adversarial training.
The state of the art in image-to-image translation has since progressed.
VQGAN, a transformer model, has been shown to outperform GANs and direct regression models \cite{esser2021vqgan} regarding standard benchmarking tasks (e.g. style transfer).
The very recently developed diffusion models \cite{li2023bbdm} show even higher accuracy compared to transformers on the same benchmark datasets.

So far, research on 6D object pose estimation has focused on finding the ideal intermediate representation, e.g. uv coordinates \cite{zakharov2019dpod}, NOCS \cite{wang2019normalized} or binary encoding \cite{su2022zebrapose}.
Simultaneously, little attention has been paid to the progression of the state of the art in image-to-image translation.
Therefore, we want to investigate novel image-to-image translation model architectures and compare them to the established GAN architecture, while building upon established forms of 2D-3D dense correspondences.

\section{Method}
\label{Method}
\begin{figure}[h!]
   \centering
   \includegraphics[width=1\linewidth]{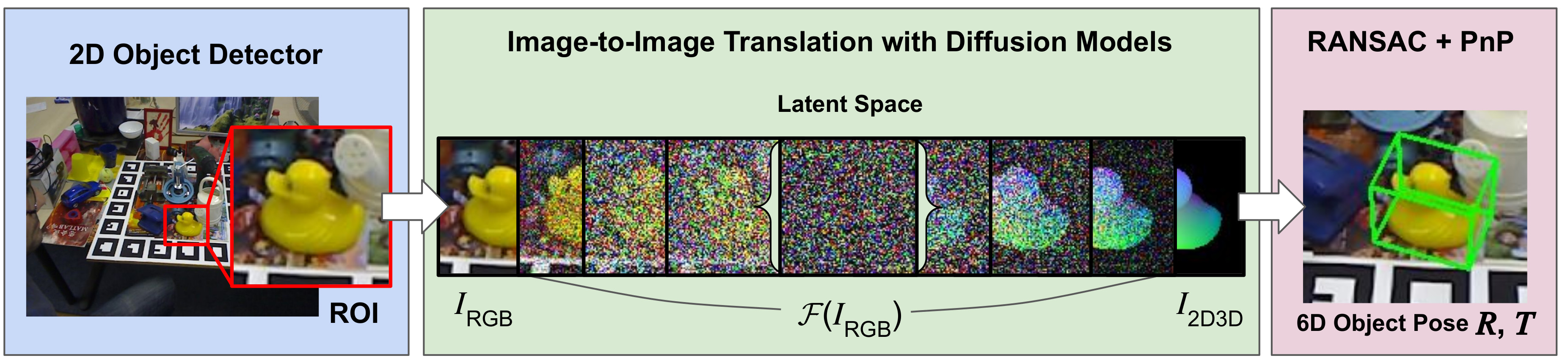}
   \caption{\textbf{Illustration of 2D-3D Dense Correspondences Estimation using Diffusion Models.} An arbitrary object detector is used to estimate a region of interest of an object. The diffusion model estimates the normalized object coordinates map from the RGB input crop. The RANSAC + PnP step solves the downstream task of 6D object pose estimation.}
   \label{fig:pipeline}
\end{figure}

Figure \ref{fig:pipeline} illustrates the full pose estimation pipeline from RGB image to 6D object pose.
To retrieve a 2D location prior, the region of interest (ROI) is cropped from the RGB image, with a 2D object detector.
The ROI is the input to the image-to-image translation model.
The image-to-image translation model learns to estimate the 2D-3D dense correspondences from the RGB crops.
The 2D object detector crops the object with a rectangular bounding box.
Therefore, the object is not entirely cropped and background pixels remain.
The learning objective of the image-to-image translation network is therefore twofold.
The networks main objective is to learn to estimate the 2D-3D dense correspondences.
Simultaneously, the network needs to learn implicitly hot to segment the object from the background.
The RANSAC + PnP step estimates the 6D object pose from the dense correspondences map.

\begin{figure}[h!]
   \centering
   \includegraphics[width=1\linewidth]{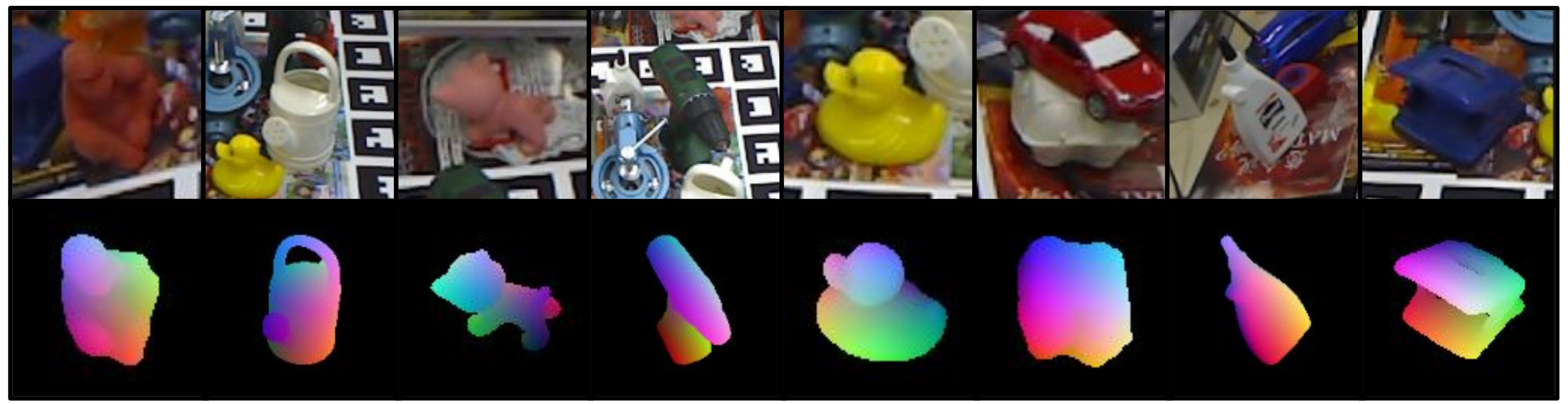}
   \caption{\textbf{Illustration of location priors (top row) and rendered dense correspondences maps (bottom row).} Ground truth translation, rotation and camera intrinsics are used to render the dense correspondences maps as identical image pair to the RGB location prior.}
   \label{fig:xyz_examples}
\end{figure}

To generate the training data for the image-to-image translation task, the object meshes are normalized, to fit into a dimensionless 1x1x1 cube.
The vertices of the object meshes are then colored with RGB values according to the XYZ positions of the vertices in the normalized object coordinate space \cite{wang2019normalized}.
The normalized and colored meshes are then rendered with the ground truth translation, rotation, and camera intrinsics.
Figure \ref{fig:xyz_examples} illustrates example location priors and dense correspondences map renderings in arbitrary poses with the objects from the Linemod-Occluded dataset.
Each object instance in the training set is re-rendered as a dense correspondences map, resulting in image pairs for image-to-image translation training.

\section{Experimental Setup}
\label{Experimental Setup}
In this section we explain the image-to-image translation algorithms which are compared.
We also introduce the dataset we use for evaluation, as well as the evaluation metrics

\subsection{Image-to-Image Translation Algorithms}
We compare two image-to-image translation methods, namely the GAN model Pix2Pix and the diffusion model BBDM.
The location priors and the RANSAC + PnP step are identical for both methods, only the image-to-image translation function $I_{DC} = \mathcal{F}(I_{RGB})$ is approximated differently.
Both models are trained under identical conditions.
Firstly, models are trained without any data augmentation, except for resizing the ROI crops to 128x128 pixels, which is the input and output size of both models.
In a second training run, both models are trained with identical data augmentation parameters, shown in Table \ref{tab:data_augmentation}.
For each run, both models are trained for 40 epochs each.

\begin{table}[h]
    \caption{\textbf{Data augmentation.} Probabilities, functions, and parameters, as introduced in \cite{wang2021gdrnet}}
    \label{tab:data_augmentation}
    \centering
    \begin{tabular}{lll}
        \toprule
        \textbf{Probability} & \textbf{Function} & \textbf{Parameters} \\
        \midrule
        0.5 & Coarse Dropout & p=0.2, size\_percent=0.05 \\
        0.5 & Gaussian Blur & 1.2*np.random.rand() \\
        0.5 & Add & (-25, 25), per\_channel=0.3 \\
        0.3 & Invert & 0.2, per\_channel=True \\
        0.5 & Multiply & (0.6, 1.4), per\_channel=0.5 \\
        0.5 & Multiply & (0.6, 1.4) \\
        0.5 & Linear Contrast & (0.5, 2.2), per\_channel=0.3 \\
        \bottomrule
    \end{tabular}
\end{table}

\subsection{Dataset}
We evaluate the image-to-image translation models on the LMO dataset \cite{brachmann2014linemodoccluded}.
It features 8 household objects sampled in randomized domains~\cite{tobin2017domainranomization} and 50 000 synthetically rendered images~\cite{denninger2019blenderproc}.
These synthetically rendered images are used for training only.
For evaluation, 1214 real-world test images are used.

\subsection{Location Priors}
Two sets of pre-computed location priors are used for object cropping.
We use the YOLOx detection results from the benchmark on object pose estimation (BOP) challenge 2023 \cite{sundermeyer2023bop} for evaluating the downstream task of pose estimation.
To evaluate object segmentation, ground truth location priors are used.


\subsection{Evaluation Metrics}
We evaluate the quality of the estimated 6D poses, as well as the the 2D-3D dense correspondences maps and the object segmentation.
The 6D object poses are evaluated using the ADD(-S) score.
ADD(-S) refers to the average distance between model points $m$, for $k_m d \geq m$.
The error threshold $k_m$ is defined with 10\%.
The calculation of $m$ is shown in Equation \ref{equ:add}.
$\overline{\mathbf{R}}$ and $\overline{\mathbf{T}}$ denote the ground truth rotation and translation, whereas $\widehat{\mathbf{R}}$ and $\widehat{\mathbf{T}}$ denote the estimated rotation and translation, while $\mathbf{x}$ denotes a model point from the model $M$.

\begin{equation}
\label{equ:add}
    m = \underset{\mathbf{x} \in M}{\text{avg}}
\begin{cases}
    \left\| (\overline{\mathbf{R}}\mathbf{x} + \overline{\mathbf{T}}) - (\widehat{\mathbf{R}}\mathbf{x} + \widehat{\mathbf{T}}) \right\| & \text{non sym.} \\
    \underset{\mathbf{x}_2 \in M}{\text{min}} \left\| (\overline{\mathbf{R}}\mathbf{x} + \overline{\mathbf{T}}) - (\widehat{\mathbf{R}}\mathbf{x}_2 + \widehat{\mathbf{T}}) \right\| & \text{sym.}
\end{cases}
\end{equation}

To compare the pose estimation results with other methods we rely on the average recall as calculated for the benchmark on object pose estimation (BOP) challenge~\cite{hodan2018bop}.
This AR score is the mean of the average recalls of visible surface discrepancy (VSD)~\cite{hodan2018bop}, maximum symmetry-aware surface distance (MSSD)~\cite{drost2017itodd}, and maximum symmetry-aware projection distance (MSPD)~\cite{hodan2020bop}.

To disentangle the image-to-image translation task from the downstream task of pose estimation, we evaluate the reconstruction loss of the 2D-3D dense correspondences maps, using the mean squared error (MSE) metric.
Equation \ref{equ:mse} shows the calculation of mean squared error, where $N$ denotes the number of pixels in the estimated and ground truth images $\widehat{I}_i$ and $\overline{I}_i$.

\begin{equation}
\label{equ:mse}
    \text{MSE} = \frac{1}{N} \sum_{i=1}^{N} (\widehat{I}_i - \overline{I}_i)^2
\end{equation}

The image-to-image task involves the implicit learning of object segmentation, to seperate the object from the background.
Therefore, we also evaluate the intersection over union (IoU) between the ground truth mask $\overline{M}_i$ and estimated masks $\widehat{M}_i$, with the number of pixels $N$, as shown in Equation \ref{equ:iou}.
The mean IoU is a measure of how well the estimated mask and the ground truth mask overlap.

\begin{equation}
\label{equ:iou}
    \text{IoU} = \frac{\sum_{i=1}^{N} \mathbf{1}_{(\widehat{M}_i \land \overline{M}_i)}}{\sum_{i=1}^{N} \mathbf{1}_{(\widehat{M}_i \lor \overline{M}_i)}}
\end{equation}

\section{Results}
\label{Results}
In this section we show results of the comparative study between Pix2Pix and BBDM regarding 6D object pose estimation accuracy, reconstruction loss and object segmentation.
We also compare with other 6D object pose estimation methods, that use 2D-3D dense correspondences and RANSAC + PnP as well.

Table \ref{tab:scores_objects} shows the main results of the comparative study between Pix2Pix and BBDM, by reporting the ADD(-S) score.
Overall, the performance of BBDM is superior to Pix2Pix, with and without augmentation.
For the objects ape and holepuncher, Pix2Pix shows better performance than BBDM, while BBDM is superior for all remaining objects.
The online augmentations influence Pix2Pix and BBDM differently.
While BBDM profits from online augmentation, the accuracy of Pix2Pix declines.
This can be attributed to the different learning objectives of the diffusion model (BBDM) and the GAN (Pix2Pix).

\begin{table}[h!]
\centering
\caption{\textbf{Main results on LMO.} ADD(-S) scores for LMO with Pix2Pix and BBDM, without (wo) and with (w) augmentations and YOLOx location priors}
\label{tab:scores_objects}
\begin{tabular}{l|cc|cc}
\toprule
\textbf{Object Name} & \multicolumn{2}{c}{\textbf{Pix2Pix}} & \multicolumn{2}{c}{\textbf{BBDM}} \\
\cmidrule(lr){2-3} \cmidrule(lr){4-5}
& \textbf{wo aug} & \textbf{w aug} & \textbf{wo aug} & \textbf{w aug} \\
\midrule
ape & 13.10 & 10.90 & 5.10 & 8.00 \\
can & 9.50 & 4.50 & 14.60 & 17.10 \\
cat & 7.00 & 7.60 & 3.50 & 11.70 \\
driller & 23.00 & 11.00 & 11.00 & 19.50\\
duck & 12.80 & 8.90 & 10.60 & 16.10 \\
eggbox & 3.90 & 3.90 & 16.10 & 20.00 \\
glue & 10.00 & 11.40 & 21.40 & 32.90 \\
holepuncher & 11.00 & 22.00 & 17.50 & 17.50 \\
\midrule
mean & 11.49 & 10.10 & 12.39 & \textbf{17.51} \\
\bottomrule
\end{tabular}
\end{table}

Table \ref{tab:mse_iou} shows the results for the mean squared error, representing the reconstruction loss, and the mean intersection over union scores, describing the accuracy of the segmentation.
BBDM is superior in terms of MSE as well as mean IoU, indicating that the diffusion model performs reconstruction of the 2D-3D dense correspondences values and segmentation of object with higher accuracy. 

\begin{table}[h!]
\centering
\caption{\textbf{Quality of the 2D-3D dense correspondences maps.} MSE and mean IoU scores for LMO, without (wo) and with (w) augmentations and ground truth location priors}
\label{tab:mse_iou}
\begin{tabular}{cc|cc|cc|cc}
\toprule
\multicolumn{4}{c}{\textbf{MSE}} & \multicolumn{4}{c}{\textbf{Mean IoU}} \\
\cmidrule(lr){1-4} \cmidrule(lr){5-8}
\multicolumn{2}{c}{\textbf{Pix2Pix}} & \multicolumn{2}{c}{\textbf{BBDM}} & \multicolumn{2}{c}{\textbf{Pix2Pix}} & \multicolumn{2}{c}{\textbf{BBDM}} \\
\textbf{wo aug} & \textbf{w aug} & \textbf{wo aug} & \textbf{w aug} & \textbf{wo aug} & \textbf{w aug} & \textbf{wo aug} & \textbf{w aug} \\
\midrule
0.092 & 0.114 & \textbf{0.060} & \textbf{0.060} & 0.816 & 0.756 & \textbf{0.851} & 0.840 \\
\bottomrule
\end{tabular}
\end{table}

\begin{figure}[h!]
   \centering
   \includegraphics[width=1\linewidth]{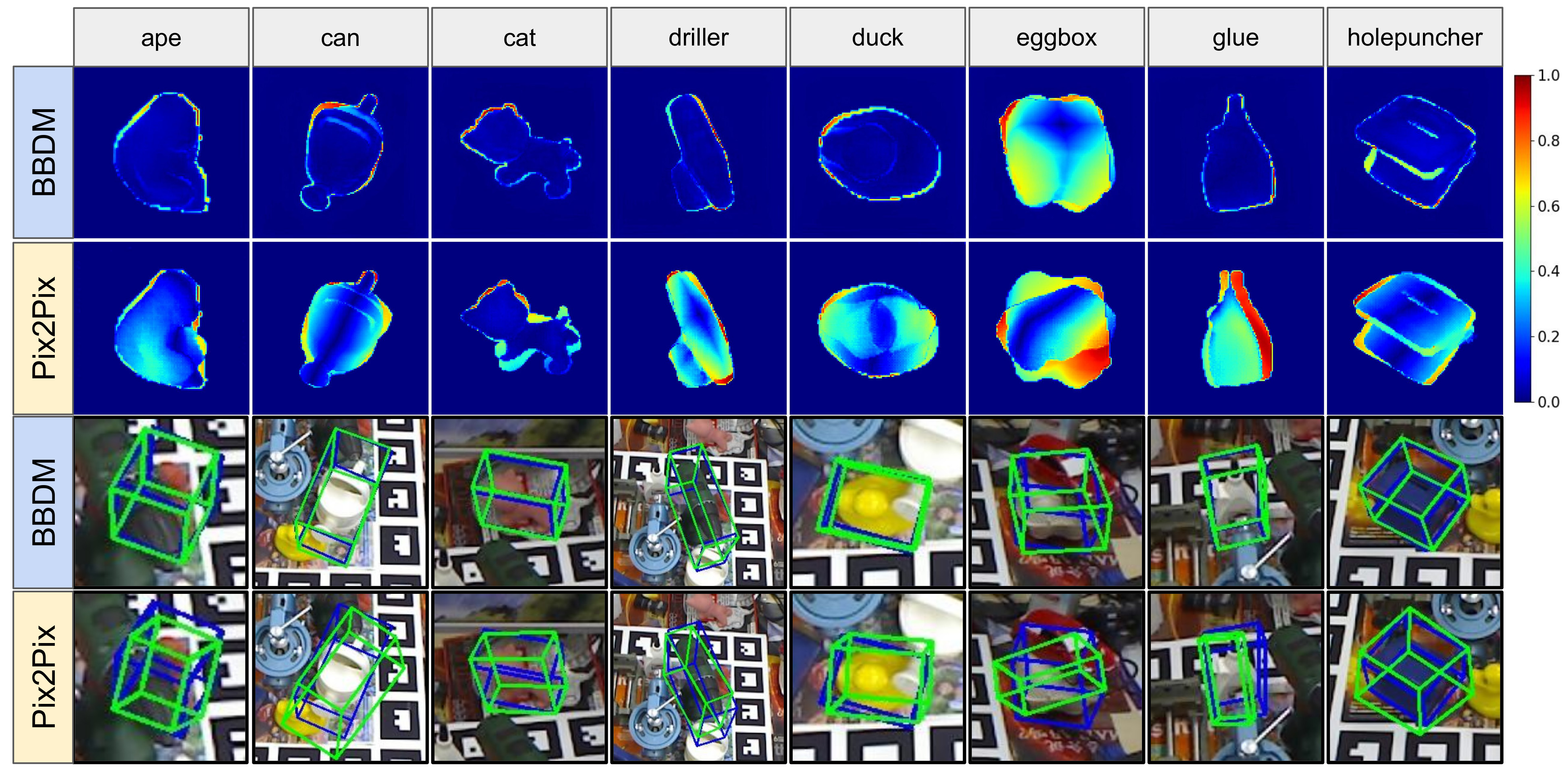}
   \caption{\textbf{Influence of reconstruction quality on estimated 6D pose.} Selected images from the LMO test data; comparison of pixel-wise error (top two rows), illustrated with the jet colormap and the influence on the estimated 6D poses (lower two rows); estimated bounding box in green, ground truth bounding box in blue.}
   \label{fig:diff_image}
\end{figure}

To illustrate the influence of the MSE and mean IoU from Table \ref{tab:mse_iou}, Figure \ref{fig:diff_image} shows selected examples from the LMO test set.
2D-3D dense correspondences are estimated by BBDM and Pix2Pix.
Absolute errors are calculated between estimated and ground truth correspondences maps.
The three channels from RGB images are averaged to a single channel error map, shown in the top two rows of Figure \ref{fig:diff_image}.
The lower two rows show the influence of the dense correspondences errors on the 6D object pose.
The selected examples in Figure \ref{fig:diff_image} demonstrate how the BBDM and Pix2Pix show errors along the object boundaries.
However, BBDM appears to draw the object boundaries more accurately and generates smoother object surfaces.
Pix2Pix shows more inaccurate object surfaces and introduces artifacts.
This is likely due to convolution and de-convolution operations, which change the size of the feature maps, in the GAN architecture of Pix2Pix.
BBDMs diffusion principle preserves feature map sizes throughout the network, maintaining sharpness of estimated images.

Table \ref{tab:time} shows the runtimes of the full 6D object pose estimation pipelines and their subparts, with either Pix2Pix or BBDM for dense correspondences estimation.
Experiments are conducted on a AMD Ryzen 9 5900X CPU and a RTX 3090 GPU. 
Pix2Pix is $\approx$10 times faster than BBDM, with a mean runtime of 0.005 seconds.
The full pipeline with BBDM takes 0.192 seconds for a single object instance.

\begin{table}[h!]
\centering
\caption{\textbf{Comparison of runtimes.} List of runtimes for each step of the 6D object pose estimation pipeline, including YOLOx detection, dense correspondences estimation and RANSAC + PnP step.}
\label{tab:time}
\begin{tabular}{@{}lcc@{}}
\toprule
 Step & \textbf{Pix2Pix} & \textbf{BBDM} \\ 
\midrule
Detection (YOLOx) & 0.081 & 0.081 \\
2D-3D Dense Correspondences & 0.005 & 0.061 \\
RANSAC + PnP & 0.050 & 0.050 \\
\midrule
Sum & \textbf{0.136} & 0.192 \\
\bottomrule
\end{tabular}
\end{table}

We compare our results with 6D object pose estimation models, with similar architecture than ours, namely Pix2Pose \cite{park2019pix2pose} and DPOD \cite{zakharov2019dpod}\footnote{Performance scores of Pix2Pose and DPOD according to the leaderboard of the BOP challenge: \url{https://bop.felk.cvut.cz/leaderboards/}}.
Both of these methods rely on 2D-3D dense correspondences as well as a RANSAC + PnP step.
As shown in Table \ref{tab:ar_scores}, our BBDM implementation outperforms Pix2Pose and DPOD.

\begin{table}[h!]
\centering
\caption{\textbf{Comparison with the State of the Art.} AR scores for LMO with Pix2Pix and BBDM, without (wo) and with (w) augmentations and YOLOx location priors; comparison with Pix2Pose \cite{isola2017pix2pix} and DPOD \cite{zakharov2019dpod}}
\label{tab:ar_scores}
\begin{tabular}{cc|cc}
\toprule
\textbf{Pix2Pix} & \textbf{BBDM} & \textbf{Pix2Pose} & \textbf{DPOD}\\
\midrule
30.03 & \textbf{40.63} & 36.30 & 16.90\\
\bottomrule
\end{tabular}
\end{table}

\section{Conclusion}
\label{Conclusion}
This paper presents a comparative study for 2D-3D dense correspondences estimation with image-to-image translation networks.
We compared the GAN model Pix2Pix and the diffusion model BBDM under identical training conditions.
Our experiments revealed that the diffusion model is superior to the GAN regarding the quality of the estimated 2D-3D dense correspondences and the accuracy of the downstream task of 6D object pose estimation.
The diffusion model segments the object from the background more accurately with sharper boundaries and estimates smooth object surfaces as compared to the GAN, which struggles with pattern-like artifacts and object boundaries lacking sharpness.
Further studies will expand this work by implementing a diffusion-based object pose estimation pipeline with improved PnP step and more resilience toward symmetries.

{
\small

{\small
\bibliographystyle{plain}
\bibliography{airov24}
}


\end{document}